\documentclass[letterpaper]{article} 
\usepackage{aaai24}  
\usepackage{times}  
\usepackage{helvet}  
\usepackage{courier}  
\usepackage[hyphens]{url}  
\usepackage{graphicx} 
\usepackage{natbib}  
\usepackage{caption} 
\usepackage{algorithm}
\usepackage{algorithmic}

    \usepackage{amsmath,amssymb} 

    \usepackage{bm}
    \newcommand{\mba}{\mathbf{a}}
    \newcommand{\mbf}{\mathbf{f}}
    \newcommand{\mbk}{\mathbf{k}}
    \newcommand{\mbp}{\mathbf{p}}
    \newcommand{\mbu}{\mathbf{u}}
    \newcommand{\mbv}{\mathbf{v}}
    \newcommand{\mbx}{\mathbf{x}}
    \newcommand{\mby}{\mathbf{y}}

    \newcommand{\mbI}{\mathbf{I}}
    \newcommand{\mbK}{\mathbf{K}}
    \newcommand{\mbQ}{\mathbf{Q}}
    \newcommand{\mbY}{\mathbf{Y}}

    \newcommand{\calN}{\mathcal{N}}
    \newcommand{\calO}{\mathcal{O}}
    \newcommand{\calX}{\mathcal{X}}
    \newcommand{\calY}{\mathcal{Y}}
    \def\argmax{\mathop{\rm arg\,max}\limits}

    \def\R{\mathbb{R}}

    \makeatletter
    \usepackage{xspace}
    \DeclareRobustCommand\onedot{\futurelet\@let@token\@onedot}
    \def\@onedot{\ifx\@let@token.\else.\null\fi\xspace}
    \def\eg{{e.g}\onedot}
    
    \def\ie{{i.e}\onedot}

    \def\etal{{et al}\onedot}
    \makeatother

\usepackage{newfloat}
\usepackage{listings}
\DeclareCaptionStyle{ruled}{labelfont=normalfont,labelsep=colon,strut=off} 
\floatstyle{ruled}
\newfloat{listing}{tb}{lst}{}
\floatname{listing}{Listing}
\title{Active Learning Guided by Efficient Surrogate Learners}
\author{
    Yunpyo~An\equalcontrib\textsuperscript{\rm 1}, Suyeong~Park\equalcontrib\textsuperscript{\rm 1}, Kwang~In~Kim\textsuperscript{\rm 2}
}
\affiliations{
    \textsuperscript{\rm 1}UNIST,     \textsuperscript{\rm 2}POSTECH\\

    \textsuperscript{\rm 1}\{anyunpyo,suyeong\}@unist.ac.kr, 
    \textsuperscript{\rm 2}kimkin@postech.ac.kr
}

\usepackage{bibentry}

\begin{document}

\maketitle

\begin{abstract}
Re-training a deep learning model each time a single data point receives a new label is impractical due to the inherent complexity of the training process. Consequently, existing active learning (AL) algorithms tend to adopt a batch-based approach where, during each AL iteration, a set of data points is collectively chosen for annotation. However, this strategy frequently leads to redundant sampling, ultimately eroding the efficacy of the labeling procedure. In this paper, we introduce a new AL algorithm that harnesses the power of a Gaussian process surrogate in conjunction with the neural network principal learner. Our proposed model adeptly updates the surrogate learner for every new data instance, enabling it to emulate and capitalize on the continuous learning dynamics of the neural network without necessitating a complete retraining of the principal model for each individual label. Experiments on four benchmark datasets demonstrate that this approach yields significant enhancements, either rivaling or aligning with the performance of state-of-the-art techniques.
\end{abstract}

\section{Introduction}
The success of deep learning heavily relies on a substantial amount of labeled data. However, creating extensive datasets poses challenges, particularly for problems demanding significant labeling efforts. As a result, the utilization of deep learning is constrained to domains with feasible labeling resources. Active learning (AL) seeks to surpass this constraint by strategically selecting the most informative data instances for labeling within a predetermined labeling budget~\cite{RXC20, Set09, AKG15}. 

Typically, in AL, a baseline learner is initially provided with a dataset where only a small or no subset is labeled. Throughout the learning process, AL algorithms analyze the data distribution and the learner's progress to recommend specific data instances for labeling. The effectiveness of AL hinges on accurately identifying both difficult (or \emph{uncertain}) data points, as well as those wielding substantial \emph{influence} over the learner's overall decisions once labeled. Identifying influential data requires capturing the global shape of the underlying data distribution. For this, existing AL algorithms primarily focus on achieving \emph{diversity} by populating areas where labels are sparsely sampled. This approach is particularly effective in the early stages of AL when the number of labeled instances is small and the learner's predictions are unreliable. However, its main limitation is the lack of consideration for the learner's progress. As more labels are acquired, the learner becomes a more reliable estimator of the underlying ground truth. In such cases, prioritize regions where the learner struggles could be more advantageous than covering areas of already established confidence.

To address this limitation, it is also essential to identify instances where the learner's predictions require deeper investigation. This entails selecting instances with the high degree of uncertainty in the learner's predictions.  However, in the early learning stages, the learner may not have a comprehensive understanding of the problem, and uncertainty estimates can be unreliable. 
Another significant challenge is the resulting redundancy in labeling. Given the computational intensity of training neural networks, updating the model for each new data instance is impracticable. Consequently, prevailing AL algorithms often adopt a batch-based strategy, wherein each AL iteration involves collectively choosing a set of data points for annotation. If not carefully managed, this can result in spatial aggregations where uncertainty spreads to neighboring instances. However, addressing uncertainties within these aggregations can frequently be achieved by annotating only one or a handful of instances and subsequently retraining the learner. Existing approaches have attempted to address this issue through additional data diversification techniques~\cite{NS04, SED19}.

In this paper, we introduce a novel AL algorithm that integrates these two essential objectives into a unified approach, using a single surrogate model for the neural network learner. Our algorithm operates in a batch mode, wherein the primary neural network learner is trained exclusively when a batch of labels is amassed. However, it mimics the continuous learning trajectory akin to that of the neural learner by harnessing an efficient Gaussian process (GP) learner, which is refreshed each time a new label is incorporated. By adopting well-established Bayesian techniques, our algorithm provides a rigorous framework for effectively and efficiently identifying influential points: At each stage, a new data instance is selected to maximize the resulting information gain over the entire dataset. Furthermore, our model rapidly identifies the most uncertain data points by \emph{instantaneously updating} the surrogate GP learner. When a new data point is added, the confidence of predictions for its spatial neighbors is immediately improved, eliminating the need for additional data diversification. 

In experiments on four datasets, our algorithm consistently outperforms or performs comparably to state-of-the-art approaches in both the early and later learning stages.

\section{Related Work}
\label{s:relatedwork}
Existing work on active learning has focused on identifying diverse points~\cite{SS18, ACK14, NS04}, uncertain points~\cite{YK19, TK01, YKK20, KAY19}, and their combinations~\cite{AZK20, ESY13}. Diversity-based approaches aim to efficiently cover the underlying data distribution. This can be achieved through \eg pre-clustering of unlabeled data~\cite{NS04}, predicting the impact of labeling individual instances on predictions~\cite{ACK14}, or maximizing mutual information between labeled and unlabeled instances~\cite{Guo10}. Sener and Savarese~\shortcite{SS18} introduced a core-set approach that minimizes the upper bound on generalization error, resulting in minimum coverings of data space through labeled data. Geifman and El-Yaniv~\shortcite{GE17} proposed a sequential algorithm that selects the farthest traversals from labeled instances to promote diversity. Gissin and Shalev-Shwartz~\shortcite{GS19} also enforced diversity by selecting data points that make labeled and unlabeled points indistinguishable. Sinha~\etal~\shortcite{SED19}'s variational adversarial active learning (VAAL) constructs a latent space using a variational autoencoder and an adversarial network to achieve sample diversity. While diversity-based approaches effectively select representative data points, they may not fully exploit the information acquired by the baseline learner in the task: At each round, assessing the trained learner on unlabeled data can offer insights into areas where additional labeling is desired, \eg areas close to the decision boundaries.

Uncertainty-based approaches focus on identifying areas where the learner's predictions are uncertain. Tong and Koller~\shortcite{TK01}'s method selects instances close to the decision boundary of a support vector machine learner. Maximizing the entropy of the predictive class-conditional distribution is a common approach for uncertainty-based selection~\cite{YKK20}. For Bayesian learners, Houlsby~\etal~\shortcite{HHG11} introduced the Bayesian active learning by disagreement (BALD) algorithm, which selects instances based on information gain. Kirsch~\etal~\shortcite{KAY19} extended BALD to deep learning with BatchBALD, suggesting multiple instances for labeling simultaneously. Tran~\etal~\shortcite{TTR19} further extended BALD by incorporating generative models to synthesize informative instances. Yoo and Kweon~\shortcite{YK19}'s learning loss algorithm trains a separate module to predict learner losses and selects instances with the highest predicted losses. Uncertainty-based approaches are particularly effective when the learner's predictions closely approximate the underlying ground truth. However, their performance can suffer at early active learning stages when the learner's predictions are unreliable.

Elhamifar~\etal~\shortcite{ESY13} proposed a hybrid approach that integrates label diversity and the learner's predictive uncertainty into a convex optimization problem. Zhang~\etal~\shortcite{ZLY20}'s state-relabeling adversarial active learning (SRAAL) extends VAAL to incorporate model uncertainty. This algorithm is specifically designed for learners that share their latent space with variational autoencoders. Ash~\etal~\shortcite{AZK20} developed the batch active learning by diverse gradient embeddings (BADGE) method, which balances diversity and uncertainty by analyzing the magnitude of loss gradients for candidate points and their distances to previously labeled points. Kim~\etal~\shortcite{KPK21} presented the task-aware VAAL (TA-VAAL) algorithm, which extends VAAL by incorporating predicted learner losses. Caramalau~\etal~\shortcite{CBK21} proposed a sequential graph convolutional network (GCN)-based algorithm that improves uncertainty sampling by analyzing the overall distribution of data using graph embeddings of data instances. Ash~\etal~\shortcite{AGK21} optimized the bound on maximum likelihood estimation error using Fisher information for model parameters. In our experiments, we demonstrate that our method outperforms or achieves comparable performance to the state-of-the-art BADGE, TA-VAAL, and sequential GCN algorithms.

Our algorithm shares a connection with Coleman~\etal~\shortcite{CYM20}'s proxy-based approach, which leverages a compact learner network to enhance the speed of the label selection process. Notably, our approach distinguishes itself through the incorporation of Bayesian GP surrogates. Although Coleman~\etal~\shortcite{CYM20}'s algorithm marks a significant stride forward in enhancing selection speed, it remains deficient in fulfilling the computational efficiency criteria essential for incremental labeling. For CIFAR10, this algorithm necessitates 83 hours to label a mere 1,000 data points. Also, unlike \cite{CYM20}, which exclusively relies on model uncertainty, our algorithm harnesses the power of a Bayesian learner to directly combine both the global influence of labeling and the inherent model uncertainty. 

\section{Active Learning Guided by Gaussian Process Surrogate Learners}
\label{s:method}
In traditional supervised learning, a function $f: \calX \rightarrow \calY$ is learned based on a labeled training set $T = \{(\mbx_1, \mby_1), \ldots, (\mbx_N, \mby_N)\} \subset \calX \times \calY$, sampled from the joint distribution of $\calX$ and $\calY$. In active learning (AL), however, we initially have access only to an input dataset $X = \{\mbx_1, \ldots, \mbx_N\}$. The AL algorithm is then given a budget to suggest $B$ data instances to be labeled. The output of AL is a labeled index subset $L$ of size $B$ that specifies the selected elements in $T$. In this work, we focus on classification problems, assuming that the data are one-hot encoded, with $\calY \subset \mathbb{R}^C$, where $C$ is the number of classes. The baseline learner $f$ produces probabilistic outputs, such that $\|f(\mbx)\|_1 = 1$ and $[f(\mbx)]_j \geq 0$ with $[f(\mbx)]_j$ being the $j$-th element (corresponding to the $j$-th class) of $f(\mbx)$. We will use $f$ to denote both the baseline learning algorithm and its output, forming a classification function. 

Our approach follows an incremental strategy for constructing $L$. Starting from an initial set of labeled points $L^0$, at each stage $t$, $L^t$ is expanded by adding a single label index $l^t$: $L^{t+1} = L^t \cup \{l^t\}$. This process is guided by a utility function $u: \{1, \ldots, N\} \rightarrow \mathbb{R}$ such that $l^t$ is chosen as the maximizer of $u$ among the indices in $\{1, \ldots, N\} \setminus L^t$. 

A good utility function should encompass two aspects: 1) the difficulty (or \emph{uncertainty}) associated with classifying each data instance, and 2) the \emph{influence} that labeling a data instance has on improving the classification decisions for other instances. However, realizing these objectives necessitates the ability to continuously observe how the baseline learner $f$ evolves at each stage (denoted as $f^t$, trained on $L^t$). Selecting $B$ data instances simultaneously at a single stage, based solely on their utility values, often leads to redundant aggregations of spatial neighbors. For instance, if a data instance $\mbx_i$ has the highest utility value $u(i)$, it is likely that its spatial neighbors also exhibit similarly high utility values. 
Labeling the entire spatial aggregation can be redundant. Applying this continuous retraining strategy becomes challenging when $f$ is a deep neural network (DNN) due to prohibitively high computational costs. Existing approaches circumvent this challenge by either designing utilities that are independent of the learner $f$~\cite{SS18,ACK14,NS04}, or by employing auxiliary processes to promote spatial diversity in labeled data~\cite{KAY19,NS04,SED19}. The latter often requires fine-tuning hyperparameters and heuristics to balance the selection of difficult labels with retaining diversity.

Our algorithm trains a computationally efficient surrogate learner $\hat{f}$ in parallel, which simulates the continuous learning behavior of the baseline $f$. We use a Gaussian process (GP) estimator for this purpose, allowing us to leverage well-established Bayesian inference techniques in designing and efficiently evaluating the utility function $u$. Notably, our approach does not necessitate additional mechanisms to promote label diversity. When a data instance with high utility is labeled, the surrogate $\hat{f}$ is instantly updated, leading to the corresponding suppression of utilities for its neighbors.

\vspace{0.2cm}
\noindent\textbf{Gaussian Process Surrogate Learner:\;\;}
For a given budget $B$, the DNN learner $f$ is trained only at every $I$-th stage. To capture the behavior of $f$ between these stages, we employ a continuously updated surrogate GP learner $\hat{f}$. At each $I$-th stage, our surrogate $\hat{f}$ is initialized to match $f$ (with softmax applied to the output layer). Between these $I$-th stages, $\hat{f}$ undergoes training using the accumulated labels. Specifically, for each newly selected data instance $\mbx$ to be labeled, the training label for the GP is computed as $\mby$ and subsequently, the prediction from $f(\mbx)$ is subtracted. We will use two types of utility functions to represent the uncertainty and influence of the predictions made by $f$ and $\hat{f}$.

Suppose that $t$ data instances have already been labeled at stage $t$. Without loss of generality, we consider these labeled points to correspond to the first $t$ points in $X$: $T^t = \{(\mbx_i, \mby_i)\}_{i=1}^t$ forms the labeled training set, while $U^t = \{\mbx_i\}_{i=t+1}^{N}$ represents the unlabeled set, \ie $L^t = \{1, \ldots, t\}$.

\vspace{0.2cm}
\noindent\emph{Kernels:\;\;}
Our GP prior is constructed by combining Gaussian kernels based on the inputs $\mbx$ and $\mbx'$, as well as the corresponding outputs of the latest learner $f(\mbx)$ and $f(\mbx')$: 
\begin{align}
\label{e:kernel}
k&(\mbx,\mbx',f(\mbx),f(\mbx'))=k_\mbx(\mbx,\mbx')k_\mby(f(\mbx),f(\mbx')),\\
k&_\mbx(\mbx,\mbx') = \overline{k}(\mbx,\mbx',\sigma_\mbx^2), k_\mby(\mby,\mby') = \overline{k}(\mby,\mby',\sigma_f^2),\nonumber\\
\overline{k}&(\mba,\mba',b)=\exp\left(-\frac{\|\mba-\mba'\|^2}{b}\right),\nonumber
\end{align}
where $\sigma_{\mbx}^2$ and $\sigma_f^2$ are hyperparameters controlling the kernel widths. The output kernel $k_\mby$ values are calculated using the stored learner values ${f(\mbx_i)}$ at the beginning of each training interval of size $I$ (see Algorithm~\ref{a:mainalg} and Sec.~\ref{s:experiments}). The use of this product kernel allows the resulting surrogate $\hat{f}$ to accurately capture the behavior of the underlying deep learner $f$, while preserving the class boundaries formed by $f$ without excessive smoothing. When training a separate baseline network with 4,500 labeled samples (for the FashionMNIST dataset; see Sec.~\ref{s:experiments} for more details), experiments conducted on 5 different random initializations showed that using the combined kernel $k(\mbx,\mbx',f(\mbx),f(\mbx'))$ reduced the mean absolute deviation between $\hat{f}$ and $f$ on the training set by 34\%, compared to using only the standard input kernel $k_{\mbx}(\mbx,\mbx')$.

\vspace{0.2cm}
\noindent\emph{Predictive Model:\;\;}
Our GP learner $\hat{f}$ employs an i.i.d. Gaussian likelihood across all classes and data instances~\cite{RasWill06}: $[\mby]_j \sim \calN([f(\mbx)]_j,\sigma^2)$, where $\calN(\mu,\sigma^2)$ represents a Gaussian distribution with mean $\mu$ and variance $\sigma^2$. By combining this likelihood with the GP prior (Eq.~\ref{e:kernel}), the prediction of $\hat{f}$ for an unlabeled point $\mbx_i \in U^t$ is represented as an isotropic Gaussian random vector of size $C$:
\begin{align}
\label{e:ogpprediction}
p(\mby_i|T^t,\mbx_i)=&\calN(\bm{\mu}_i^t,\bm{\Sigma}_i^t), \text{ where }\\
\bm{\mu}_i^t=&(\mbk_i^t)^\top(\mbK^t+\sigma^2\mbI)^{-1}\mbY^t,\nonumber\\ 
\bm{\Sigma}_i^t=&(1-(\mbk_i^t)^\top(\mbK^t+\sigma^2\mbI)^{-1}\mbk_i^t)\mbI,\nonumber
\end{align}
$\mbk_i^t = [k(\mbx_i,\mbx_1,f(\mbx_i),f(\mbx_1)),\ldots,k(\mbx_i,\mbx_{t},f(\mbx_i),f(\mbx_{t}))]^\top$, $[\mbK]_{mn}^t=k(\mbx_m,\mbx_n,f(\mbx_m),f(\mbx_n))$ for $1\leq m,n\leq t$, and $\mbY^t=[\mby_1^\top,\ldots,\mby_{t}^\top]^\top$. This model requires inverting the kernel matrix $\mbK^t$ of size $t\times t$ which grows quickly as AL progresses. To ensure that the computational complexity of $\hat{f}$ predictions (Eq.~\ref{e:ogpprediction}) remains manageable, we employ a sparse GP approximation~\cite{SnelGhah06} for our product kernel $k$ using two sets of basis points $U=\{\mbu_1,\ldots,\mbu_K\}$ and $V=\{\mbv_1,\ldots,\mbv_K\}$:
\begin{align}
\label{e:spgpprediction}
\bm{\mu}_i^t&\approx(\mbk_i^t)^\top(\mbQ^t)^{-1}\mbK_{XP}^t(\bm{\Lambda}^t+\sigma^2\mbI)^{-1}\mbY^t,\\
\bm{\Sigma}_i^t&\approx (1-(\mbk_i^t)^\top (\mbK_{PP}^{-1}-(\mbQ^t)^{-1}) \mbk_i^t)\mbI+\sigma^2\mbI, \text{ where }\nonumber\\
\mbQ^t&=\mbK_{PP}+(\mbK_{XP}^t)^\top (\bm{\Lambda}^t+\sigma^2\mbI)^{-1}\mbK_{XP}^t,\nonumber
\end{align}
$\mbk_i^t = [k(\mbx_i,\mbu_1,f(\mbx_i),\mbv_1),\ldots,k(\mbx_i,\mbu_{K},f(\mbx_i), \mbv_{K})]^\top$, 
for $1\leq m\leq t$, $[\mbK_{XP}^t]_{mn}=k(\mbx_m,\mbu_n,f(\mbx_m),\mbv_n)$,  $[\mbK_{PP}]_{mn}=k(\mbu_m,\mbu_n,\mbv_m,\mbv_n)$, and $\bm{\Lambda}^t$ is a diagonal matrix with its $n$-th entry $[\bm{\Lambda}^t]_{nn}$ defined as $1-(\mbk_n^t)^\top \mbK_{PP}^{-1}\mbk_n^t$. The basis points $U$ are obtained by extracting the cluster centers of $X$ using $K$-means clustering, while $V$ is randomly sampled as $C$-dimensional probability simplices. The complexity of evaluating a prediction (Eq.~\ref{e:spgpprediction}) now scales linearly with the number of labeled data points: $\calO(tK^2)$.

\vspace{0.2cm}
\noindent\textbf{Influence-Based Utility $u^1$:\;\;}
As a Bayesian model, the output of $\hat{f}$ for an unlabeled input $\mbx_i$ is presented as a probability distribution $p(\mby_i|T^t,\mbx_i)$, which naturally quantifies uncertainty. The diagonal elements of the predictive covariance matrix for input $\mbx_i$ correspond to the entropies of the predictions for each class.\footnote{The variance of a Gaussian distribution is proportional to its entropy.} Therefore, the trace of this matrix indicates the overall uncertainty of the current model's prediction for $\mbx_i$. 

However, relying solely on individual data point entropies fails to capture their \emph{influence} when labeled, and optimizing based on this criterion alone tends to prioritize outliers. This behavior arises because (the diagonal terms of) the predictive covariance matrix $\bm{\Sigma}_i^t$ tends to increase as the corresponding unlabeled points deviate from the labeled training set: See \cite{SW05} for an analysis of this behavior for large-scale problems. For a more meaningful measure of utility, we define $u^1(i)$ based on the reduction in predictive entropies of the \emph{entire remaining unlabeled set} $U^{t-1}$ upon labeling $\mbx_i$. To achieve this, we maintain the predictive covariance of the unlabeled set at each stage, storing only a single diagonal entry per data point due to the isotropic nature of the covariance estimates in our model:
\begin{align}
u^1(i)&= \sum_{j=t+1}^N \text{trace}[\bm{\Sigma}_j^t]-\sum_{j=t+1}^N \text{trace}[\bm{\Sigma}_j(i)]\nonumber\\
&= C\sum_{j=t+1}^N (\mbk_j^t)^\top (\mbQ(i)^{-1}-(\mbQ^t)^{-1}) \mbk_j^t,\nonumber
\end{align}
where $\bm{\Sigma}_j(i)$ denotes the covariance for $\mbx_j\in U^{t-1}\setminus \{\mbx_i\}$ predicted by the model trained on $T^t$ and $\mbx_i$ (assuming that $\mbx_i$ is labeled),
\begin{align}
\label{e:qmat}
\mbQ(i) =& \mbK_{PP}\;+\\
&\begin{pmatrix}\mbK_{XP}^t\\
(\mbk_i^t)^\top\end{pmatrix}^\top\left(\begin{pmatrix}\bm{\Lambda}^t&0\\
0\lambda_i\end{pmatrix}+\sigma^2\mbI\right)^{-1}\begin{pmatrix}\mbK_{XP}^t\\
(\mbk_i^t)^\top\end{pmatrix},\nonumber
\end{align}
and $\lambda_i=1-(\mbk_i^t)^\top \mbK_{PP}^{-1}\mbk_i^t$. Directly calculating $u(i)$ requires $\calO(tK^2+K^3)$-time which is prohibitive even for moderately-sized datasets. A computationally efficient solution is obtained by observing that the difference between $\mbQ(i)$ and $\mbQ^t$ is rank one in that 
\begin{align}
\label{e:rankone}
\mbQ(i)=\mbQ^t+\mba\mba^\top \text{ with }\;\; \mba=\frac{\mbk_i^t}{\sqrt{\lambda_i+\sigma^2}}.
\end{align}
Applying the \emph{Sherman–Morrison–Woodbury} matrix identity~\cite{Sch16} to Eq.~\ref{e:qmat} using Eq.~\ref{e:rankone}, we obtain
\begin{align}
u^1(i)&= C\sum_{j=t+1}^N (\mbk_j^t)^\top\left( \frac{(\mbQ^t)^{-1}\mbk_i^t(\mbk_i^t)^\top(\mbQ^t)^{-1}}{\lambda_i+\sigma^2+(\mbk_i^t)^\top(\mbQ^t)^{-1}\mbk_i^t}\right) \mbk_j^t.\nonumber
\end{align}
With this form, the utility $u^1(i)$ can be evaluated in $\calO(K^2)$-time when $(\mbQ^t)^{-1}$ is provided. Once the inverse $(\mbQ^0)^{-1}$ is explicitly computed at stage 0, the inverse $(\mbQ^t)^{-1}$ at each subsequent stage $t$ can be calculated from $(\mbQ^{t-1})^{-1}$ in $\calO(K^2)$ time. 

\vspace{0.2cm}
\noindent\emph{Discussion:\;\;} 
Our approach draws inspiration from Bayesian AL methods that aim to minimize the entropy of the learner's parameters and their approximations(\eg \cite{HHG11,KAY19}). However, unlike these approaches, we minimize the entropy over predictions made on the entire dataset, rather than focusing on model parameters. This allows us to calculate entropy independently of the specific parametric forms of the learner, and it enables us to use a GP as a surrogate for the deep learner $f$.

Evaluating $u^1$ does not require knowing the actual label of each candidate point $\mbx_i$, even though $u^1$ was derived under the assumption that $\mbx_i$ is labeled. This property arises from the i.i.d. Gaussian noise model. In general, for an underlying classification function $\tilde{f}$, the likelihood $p(\mby|\tilde{f}(\mbx))$ of observing data $\mby$ given an input $\mbx$ is not Gaussian. In such cases, logistic likelihood models are commonly employed in GP models. However, these models yield covariance predictions that explicitly depend on the labels of each candidate point, while the labels become available only after the corresponding data points are actually selected.

Our utility $u^1$ uses the expected reduction of predictive variances when labeled. Theoretically, a more appealing approach might be to use the expected reduction of test errors. However, since ground-truth labels are not available, such error reduction cannot be directly calculated. Existing approaches therefore introduced certain model assumptions~\cite{FRD14,VK10} (which might hold for only specific learners). 

\vspace{0.2cm}
\noindent\textbf{Uncertainty-Based Utility $u^2$:\;\;}
This is based on the entropies of the class-conditional predictive distributions generated by DNNs. This differs from the entropy used in $u^1$, which is defined for continuous GP predictive distributions. A straightforward approach to design such a utility is to directly measure the entropy of $f$-prediction at each stage:
\begin{align}
\label{e:u2}
\widehat{u}^2(i) = \text{Ent}(f^t(\mbx_i)),
\end{align}
where $\text{Ent}(\mbp)$ represents the entropy of a distribution $\mbp$. 

Maximizing $\widehat{u}^2(i)$ effectively selects the most uncertain point for classification. However, implementing this strategy requires retraining $f^t$ at each stage, which is infeasible due to the high training cost. Instead, we train $f$ only at every $I$-th stage and use the Gaussian process (GP) surrogate to estimate entropy values for the intermediate stages. Let's consider stage $s$, where the DNN classifier $f^s$ is newly trained, and we calculate the corresponding entropy values $\{\text{Ent}(f^s(\mbx_{t+1})),\ldots,\text{Ent}(f^s(\mbx_N))\}$. The entropy values at intermediate stages are generated by calibrating the original DNN entropy values $\{\text{Ent}(f^s(\mbx_i))\}$ using the surrogate predictions. Initially, we set the calibrated entropy $\widetilde{\text{Ent}}^s(f(\mbx_i))$ at stage $s$ as $\text{Ent}(f^s(\mbx_i))$. The new entropy at stage $t>s$ is then calculated as follows:
\begin{align}
u^2(i) = \widetilde{\text{Ent}}^t(f(\mbx_i)) = \widetilde{\text{Ent}}^{t-1}(f(\mbx_i))\frac{\text{Ent}(\rho(\hat{f}^t(\mbx_i)))}{\text{Ent}(\rho(\hat{f}^{t-1}(\mbx_i)))},\nonumber
\end{align}
where $\rho(\hat{f}^t(\mbx_i))$ is the softmax output entropy of the mean prediction $\bm{\mu}_i^t$ made by the GP surrogate $\hat{f}^t$. Annotating a data instance $\mbx_i$ not only reduces the uncertainty of GP predictions at $\mbx_i$ but also influences its neighboring points. As a result, the calibrated entropies reflect the continuous reduction of uncertainties caused by the introduction of new labeled points during the intermediate stages.

\vspace{0.2cm}
\noindent\emph{Discussion.} 
We also investigated the possibility of introducing a utility $u^3$ as an alternative to $u^2$, based on the reduction of predictive entropies for all data instances in the unlabeled set when a candidate point $\mbx_i$ is labeled. However, since this measure involves the label of each candidate $\mbx_i$, direct evaluation is not possible. Instead, for each class $j\in\{1,\ldots,C\}$, we tentatively assigned the corresponding class label to $\mbx_i$ and computed the resulting entropy reduction. The final utility $u^3(i)$ was obtained by averaging these hypothetical entropy values, weighted by the corresponding class probabilities ${[f(\mbx_i)]_j}$ predicted by the learner. In our preliminary experiments on the FashionMNIST dataset (see Sec.~\ref{s:experiments}), we observed that the original utility $u^2$ achieved a final classification accuracy that was, on average, only 0.16\% lower than that of $u^3$, while being approximately 50 times faster. The main computational bottleneck in evaluating $u^3$ was the computation of entropy values for the entire unlabeled set for each hypothesized candidate. The competitive performance of $u^2$ can be attributed to the fact that maximizing $u^2$ does not select outliers, unlike the use of predictive entropies in $u^1$. When employing monotonically increasing activations such as sigmoid and ReLU, DNN predictions tend to be overly confident on outliers, which are points that deviate significantly from the training set. While this artifact is generally not desirable for classification purposes, it has a favorable side-effect in AL, as outliers are assigned low class-conditional entropy values and are not selected.

\begin{algorithm}[tb]
\caption{Active learning guided by GP proxies.}
{
\textbf{Input}: Input data $X$, budget $B$, training interval $I$, and initial label set $L^0$\\
\textbf{Output}: {Label index set $L^B$}
}
\begin{algorithmic}[1] 
\FOR{$t=0,\ldots,B$}
\STATE Calculate GP predictions and update $\mbQ^t$ (Eq.~\ref{e:spgpprediction});\\
\IF {$\emph{mod}(t,I)=0$}
\STATE Train $f^t$, and calculate $\{\text{Ent}(\rho(f^t(\mbx_i)))\}$ and $\{k_\mby(f^t(\mbx_i),\mbv_j)\}$ (Eq.~\ref{e:kernel}); \ENDIF
\STATE Evaluate $u^1$ and $u^2$;\\
\STATE Calculate the test accuracy estimate $P(f^t)$;\\
\STATE Generate $u$ by combining $u^1$ and $u^2$ using $P(f^t)$ (Eq.~\ref{e:ucombination});
\STATE $l^t=\argmax u$; 
\STATE $L^t=L^{t-1}\cup \{l^t\}$;
\ENDFOR
\end{algorithmic}
\label{a:mainalg}
\end{algorithm}

\vspace{0.2cm}
\noindent\textbf{Combining $u^1$ and $u^2$:\;\;}
Our utility functions, $u^1$ and $u^2$, exhibit complementary strengths. $u^1$ quantifies the overall reduction of uncertainty across the entire unlabeled set, providing an effective approach for exploring the data space, especially when the learner lacks sufficient information about the problem. It is particularly useful in scenarios where the learner's accuracy is low due to limited labeled data. However, $u^1$ is agnostic to the specific task at hand, as the predictive covariance $\bm{\Sigma}_i^t$ is solely determined by the distribution of input data instances in $X$ and remains independent of the acquired labels (Eq.~\ref{e:spgpprediction}). On the other hand, $u^2$ capitalizes on the label information captured by the entropy of $f(\mbx)$. However, in cases where the performance of the learner $f$ is limited, the estimated entropies themselves can be unreliable indicators. To combine the strengths of $u^1$ and $u^2$, we employ a convex combination:
\begin{align}
\label{e:ucombination}
u(i) = (1-P(f^t))\overline{u}^1(i)+P(f^t)\overline{u}^2(i),
\end{align}
where $\overline{u}^1$ and $\overline{u}^2$ represent the normalized versions of $u^1$ and $u^2$, respectively, based on their respective standard deviations computed on the entire dataset. Here, $P(f^t)\in[0,1]$ is an estimate of the test accuracy. As we do not have access to test data, we instead treat the newly labeled training data point at stage $t$ as a single test point before it is added to $L^{t-1}$ and accumulate the resulting accuracies from the initial AL stage $t=0$. Algorithm~\ref{a:mainalg} summarizes the proposed algorithm.

\vspace{0.2cm}
\noindent\textbf{Approximation Quality of the GP Proxies:\;\;}
By design, our GP surrogate $\hat{f}$ coincides with $f$ at every $I$-th stage, while it may exhibit deviations from $f$ in between these stages. However, empirical observations indicate that $\hat{f}$ effectively captures the behavior of $f$ thanks to the product kernel $k$ (see Eq.~\ref{e:kernel}). To validate this, we trained a separate neural network learner $f'$ at an intermediate stage with 2,500 labels (for the FashionMNIST dataset). The signal-to-noise ratio of $\hat{f}$ compared to $f'$ at $t=2,500$ was measured to be 15.49dB. This noise level is comparable to the variations observed in $f'$ resulting from retraining the DNN learners using the same training data but with random initializations.

\begin{figure*}[t]
\centering
\includegraphics[width=0.4835\linewidth]{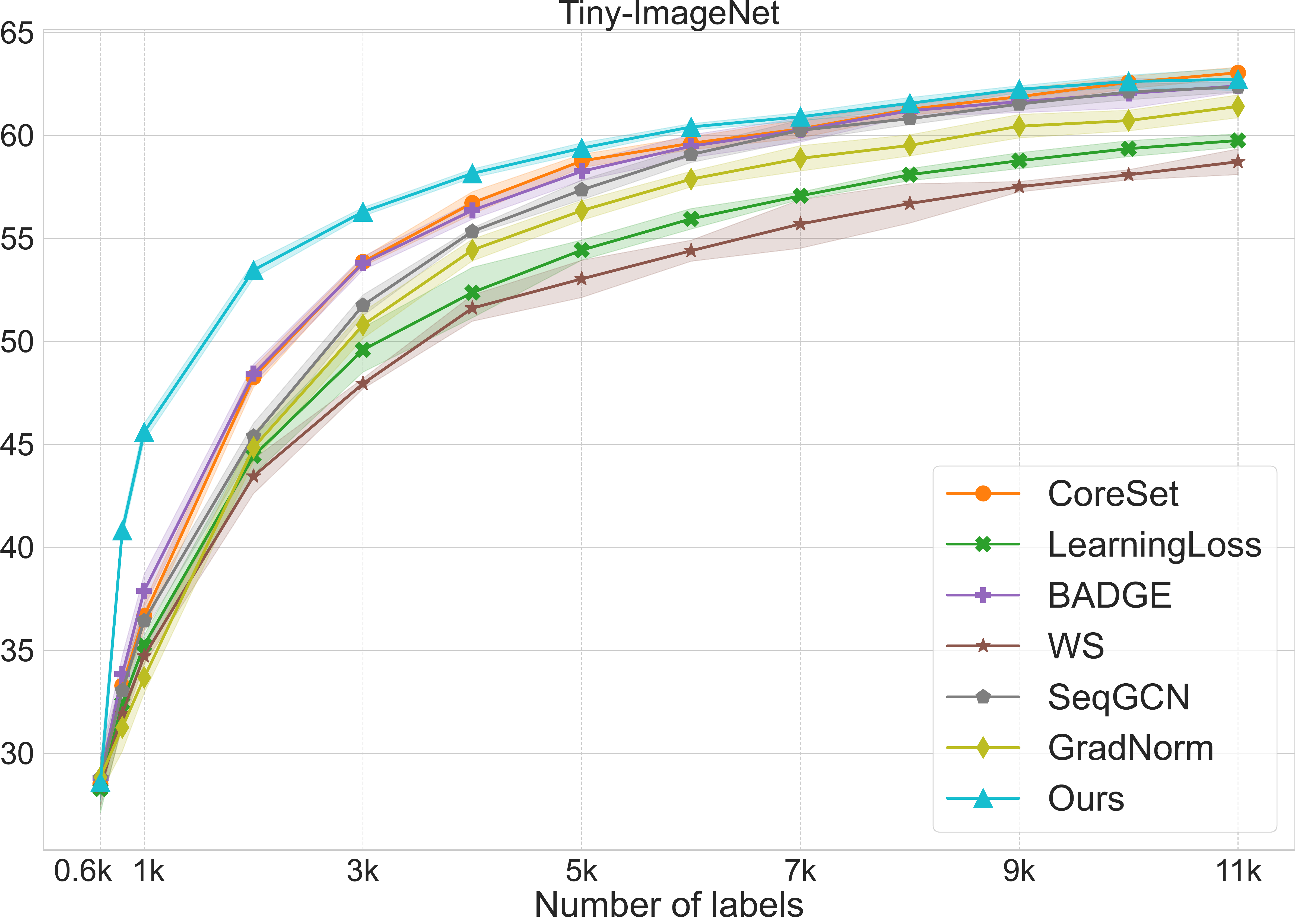}\;\;
\includegraphics[width=0.4835\linewidth]{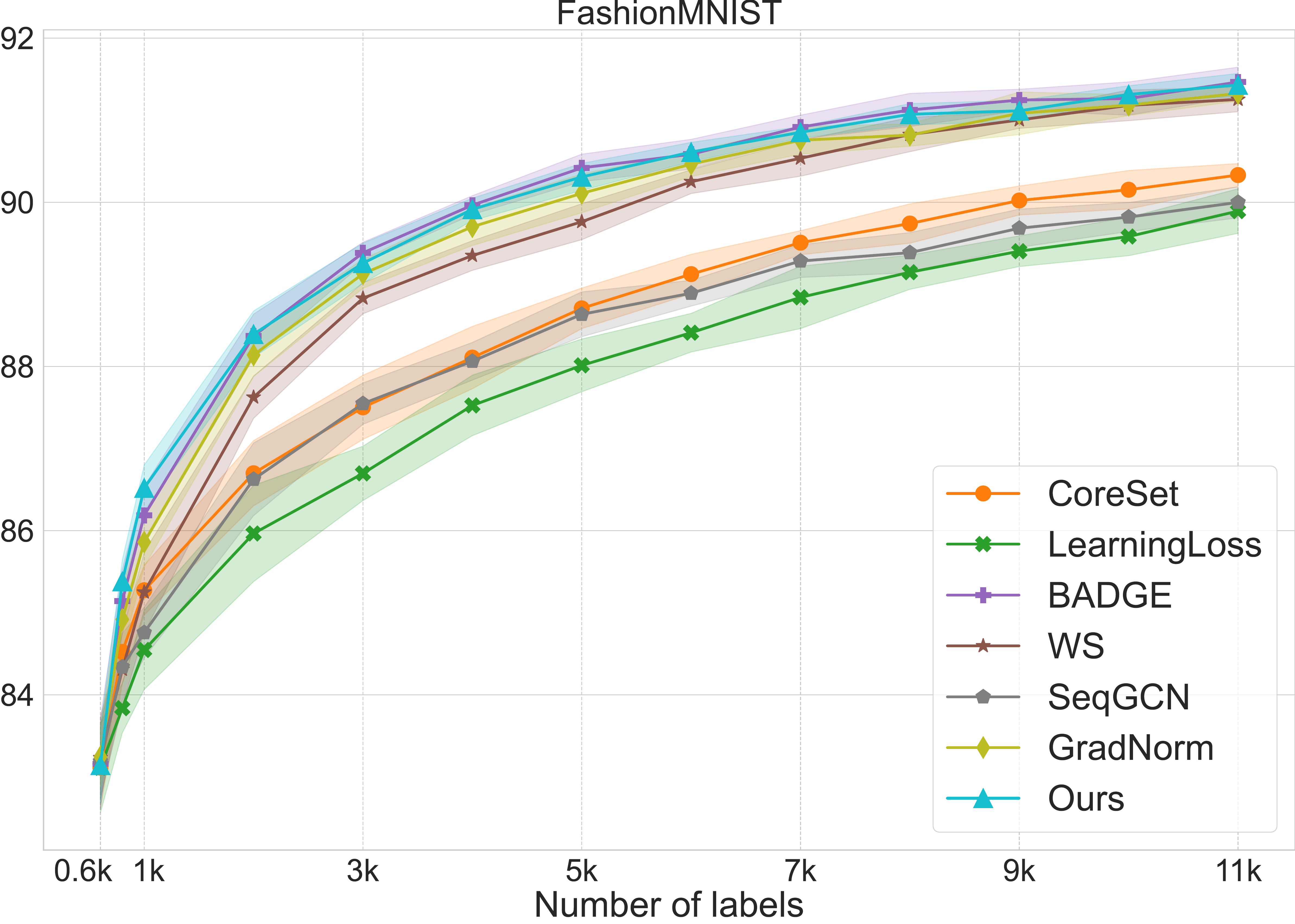}\\
\vspace{0.2cm}
\includegraphics[width=0.4835\linewidth]{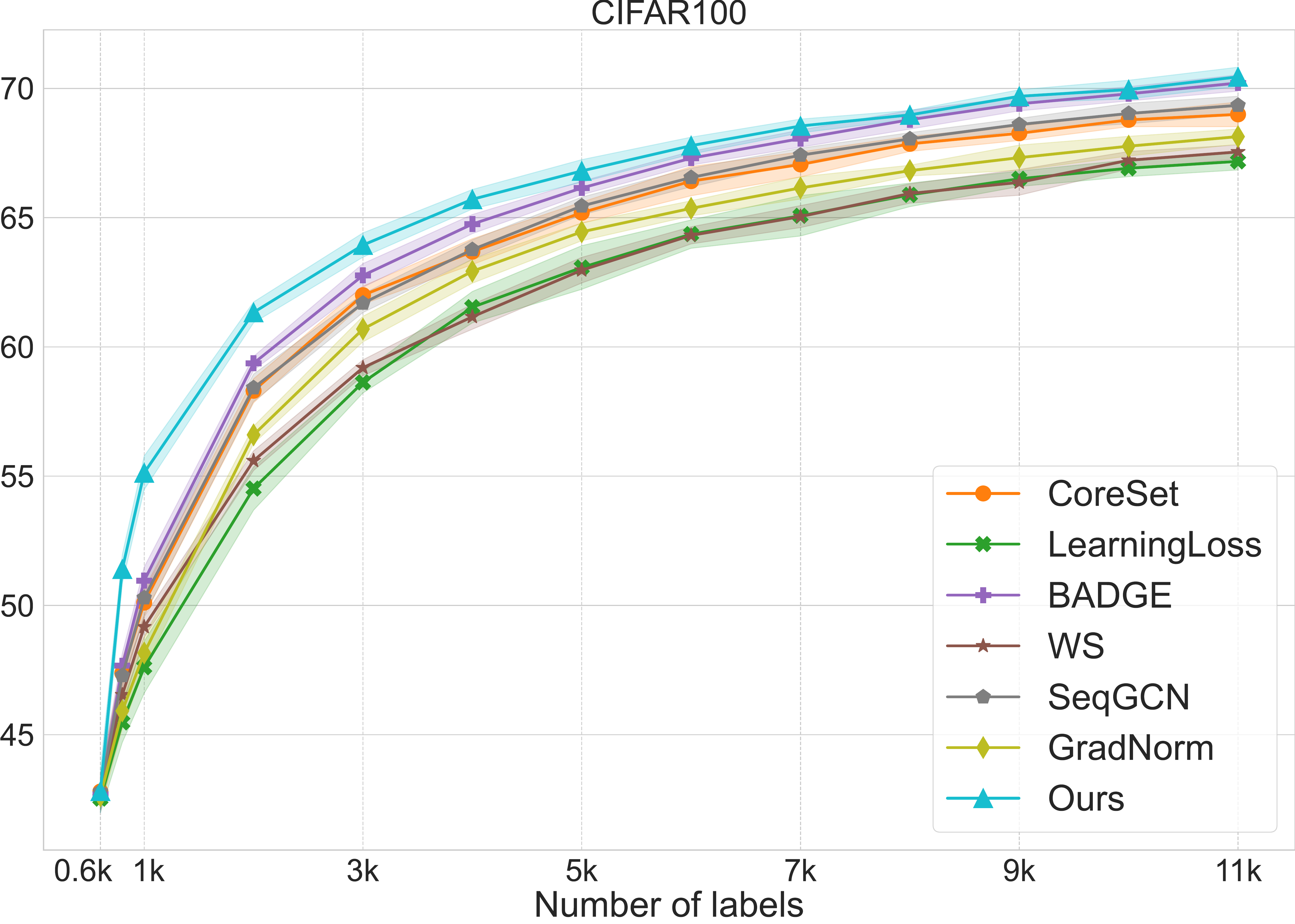}\;\;
\includegraphics[width=0.4835\linewidth]{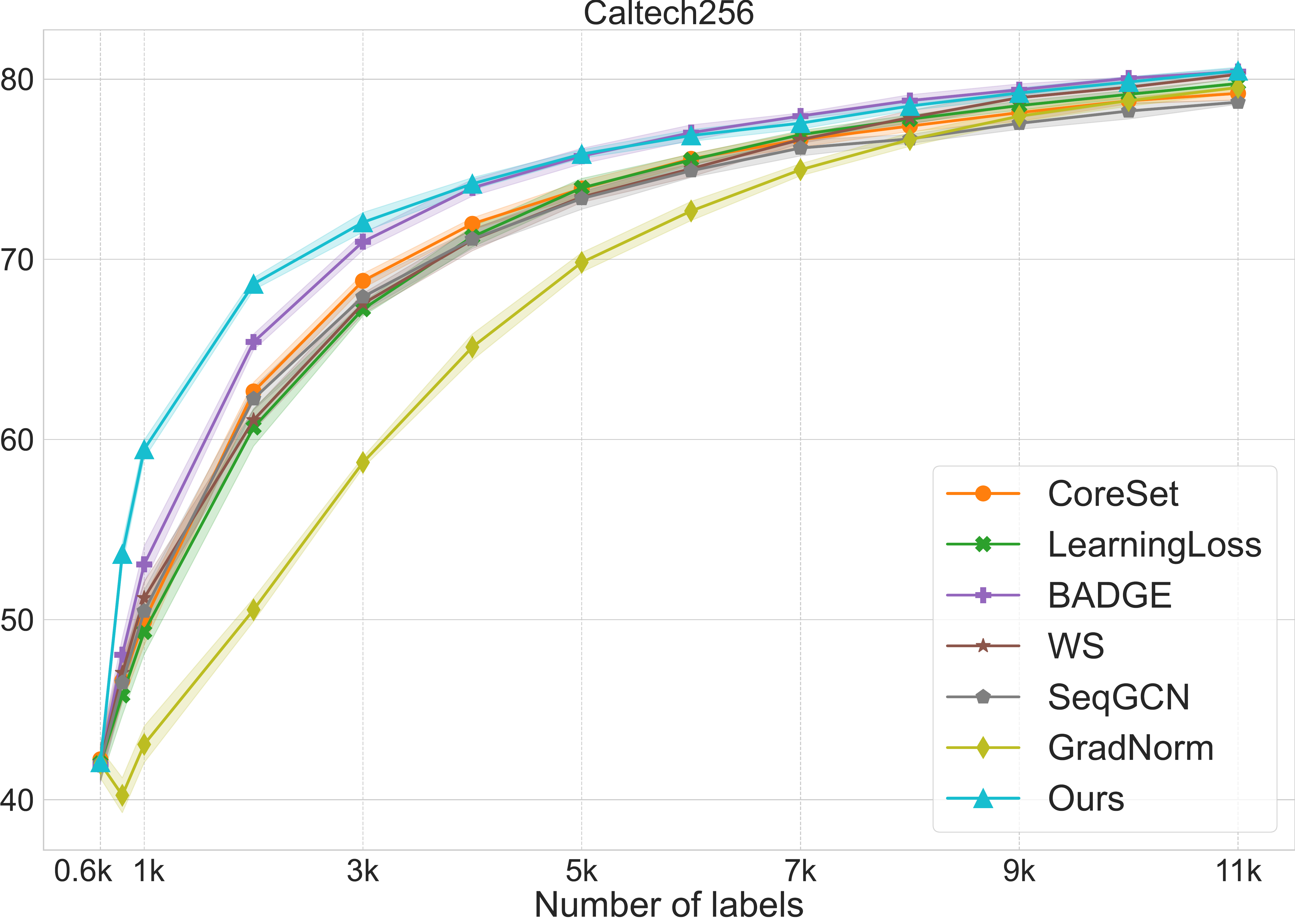}
\caption{Mean accuracy (in \%) across ten repeated experiments for different active learning algorithms including \emph{CoreSet}~\cite{SS18}, \emph{LearningLoss}~\cite{YK19}, \emph{BADGE}~\cite{AZK20}, \emph{SeqGCN}~\cite{CBK21}, \emph{GradNorm}~\cite{WLY22}, weight decay scheduling (\emph{WS})~\cite{YKK20}, and our algorithm, with random network initializations. The widths of the shaded regions represent twice the standard deviations.}
\label{f:results}
\end{figure*}

\begin{figure*}[th!]
\centering
\includegraphics[width=0.48\linewidth]{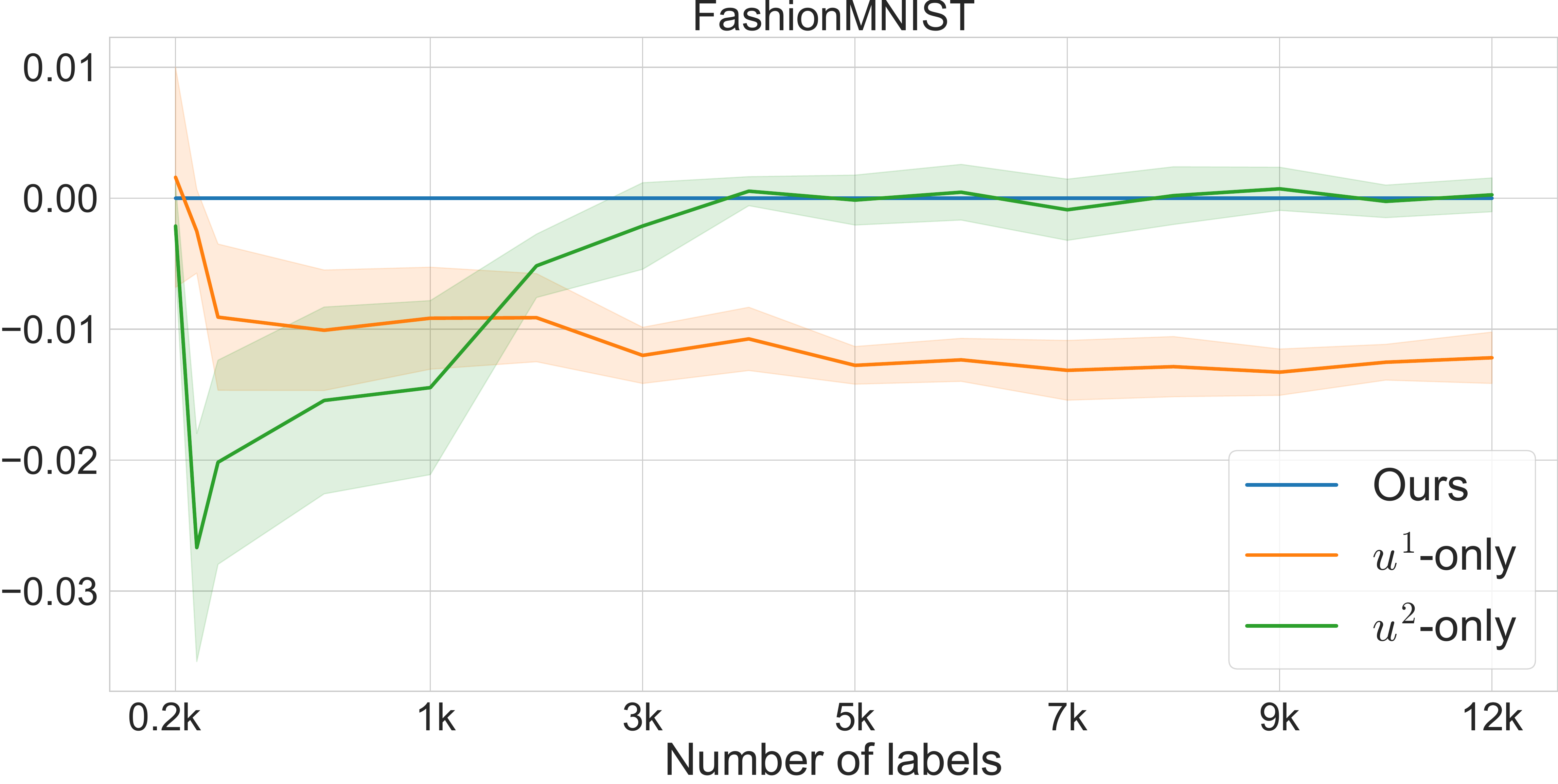}\;\;\;
\includegraphics[width=0.48\linewidth]{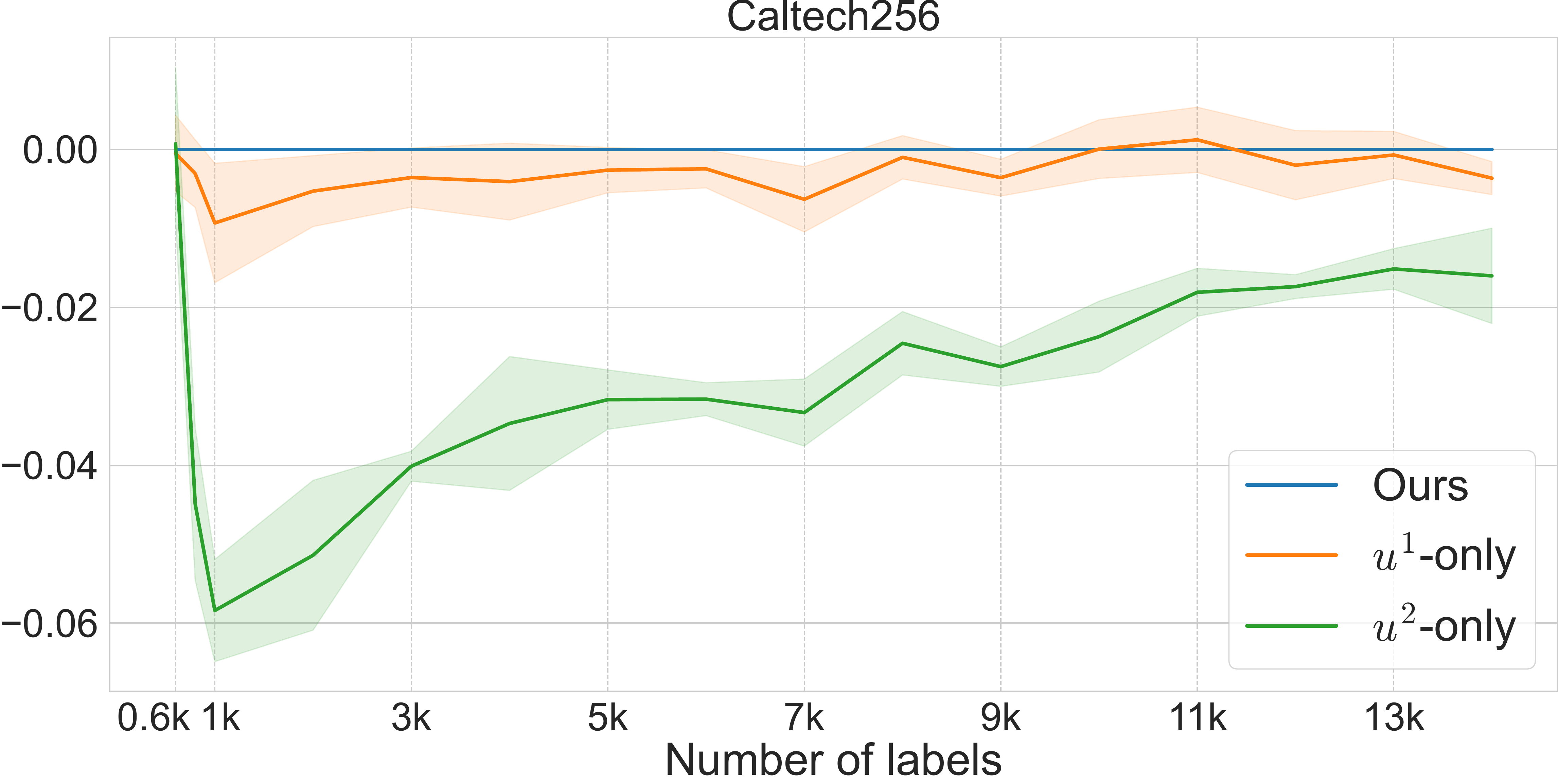}
\caption{Performance comparison of two variants of our final algorithm: \emph{$u^1$-only} and \emph{$u^2$-only}, using solely the $u^1$ and $u^2$ utilities, respectively. The y-axis represents the accuracy difference compared to our final algorithm. Negative differences indicate superior performance of the final version. Our final algorithm achieves a favorable balance between maximizing influence and reducing uncertainty by leveraging the complementary strengths of the $u^1$ and $u^2$ utilities. }
\label{f:ablation}
\end{figure*}

\vspace{0.2cm}
\noindent\textbf{Hyperparameters and Time Complexity:\;\;}
Our algorithm involves three hyperparameters. The number of basis points $K$ for $U$ and $V$ (Eq.~\ref{e:spgpprediction}) is fixed at 500, balancing between computational complexity and approximation accuracy of GP predictions. The input kernel parameter $\sigma_\mbx$ is set to 0.5 times the average distance between data instances in $X$, while the output kernel parameter $\sigma_f$ is determined as the number of classes per dataset. The noise level $\sigma^2$ (Eq.~\ref{e:ogpprediction}) is kept small at $10^{-10}$. These hyperparameters were chosen without using problem- or dataset-specific information. While fine-tuning them for each task and dataset could potentially improve performance, it would require additional validation labels, which are often scarce in AL scenarios.

The computational complexity of each iteration of our algorithm is $\calO(K^2\times N)$, where $N$ is the number of unlabeled data instances and $K$ is the rank of the sparse GP approximation (Eq.~\ref{e:spgpprediction}). On average, our algorithm takes approximately 0.35 seconds to suggest a point for labeling on the FashionMNIST dataset. In comparison, the run-times of other methods such as \emph{VAAL}, \emph{CoreSet}, \emph{LearningLoss}, \emph{BADGE}, \emph{WS}, \emph{SeqGCN}, and \emph{TA-VAAL} (see Sec.\ref{s:experiments}) were 23.73, 0.05, 0.12, 0.58, 0.03, 0.86, and 23.78 seconds, respectively. While instantiating continuous learning, our approach incurs comparable computational costs.

\section{Experiments}
\label{s:experiments}
\paragraph{Settings:} We evaluated the performance of our algorithm using four benchmark datasets: The Tiny-ImageNet dataset~\cite{LY15}, CIFAR100~\cite{Kri09}, FashionMNIST~\cite{XiaRasVol17}, and Caltech256~\cite{GriHolPer07} datasets. For comparison, we also performed experiments with Sener and Savarese~\shortcite{SS18}'s core-set approach (\emph{CoreSet}), Yoo and Kweon~\shortcite{YK19}'s learning loss (\emph{LearningLoss}), Caramalau~\etal~\shortcite{CBK21}'s sequential GCN-based algorithm (\emph{SeqGCN}), Ash~\etal~\shortcite{AZK20}'s batch AL by diverse gradient embeddings (\emph{BADGE}), Wang~\etal~\shortcite{WLY22}'s gradient norm-based approach (\emph{GradNorm}), and Yun~\etal~\shortcite{YKK20}'s weight decay scheduling scheme (\emph{WS}). In the accompanying supplementary document, we also present the results of Sinha~\etal~\shortcite{SED19}'s variational adversarial active learning (\emph{VAAL}) and the task-aware VAAL-extension proposed by Kim~\etal~\shortcite{KPK21} (\emph{TA-VAAL}). Throughout the experiments, we initiated the process by randomly selecting 600 images and labeling them, which were then used to train the baseline learner. Subsequently, the AL algorithms augmented the labeled set until it reached the final budget of $B=$11,000. To evaluate the label acquisition performance during the early stages of learning, the learners were assessed at 600, 800, and 1,000 labels, followed by evaluations at every 1,000 additional labels ($I$=1,000). These experiments encompassed a range of labeling budgets $B=\{600,800,1000,\ldots,11,000\}$.

For the baseline learner of the AL algorithms, we initially evaluated ResNet18~\cite{HeZhaRen16}, ResNet101~\cite{HeZhaRen16}, and VGG16~\cite{SimZis15}, all combined with fully connected (FC) layers matching the number of classes per dataset. Among them, we selected a ResNet101 pre-trained on ImageNet; Combining the \emph{pool5} layer of ResNet101 with three FC layers consistently outperformed the other networks. Our learners were trained using stochastic gradient descent with an initial learning rate of 0.01. The learning rate was reduced to 10\% for every 10 epochs. The mini-batch size and the total number of epochs were fixed at 30 and 100, respectively. For the GP surrogate $\hat{f}$, we used the \emph{pool5} layer outputs of ResNet101 as inputs $\mbx$. All experiments were repeated ten times with random initializations and the results were averaged. 

\vspace{0.2cm}
\noindent\textbf{Results:\;\;}
Figure~\ref{f:results} presents a summary of the results. While \emph{CoreSet} is effective in identifying diverse data points by analyzing the overall distribution of data, its performance tends to decline in later stages of learning when the baseline learner $f$ provides more reliable uncertainty estimates. A comparable behavior was demonstrated by \emph{VAAL} (in the supplementary document). This is because diversity-based methods do not directly leverage $f$'s predictions. On the other hand, \emph{WS} and \emph{LearningLoss} use this task-specific information to achieve higher accuracies compared to \emph{CoreSet} in later stages of Caltech256. 

The performance of the different algorithms exhibited notable variations across the datasets. FashionMNIST, consisting of only 10 classes, demonstrated that even with a limited number of labeled samples, the baseline learners produced highly accurate uncertainty predictions. Consequently, uncertainty-based methods, \emph{LearningLoss} and \emph{WS} showed superior performance on these datasets. Conversely, CIFAR100, Caltech256, Tiny-ImageNet, with a larger number of classes, posed challenges as the class predictions and uncertainty estimates of the learner $f$ became unreliable, even with a larger number of labeled samples. In such cases, diversity-based method, namely \emph{CoreSet} demonstrated higher accuracies.

By combining diversity and uncertainty, \emph{BADGE} achieved significantly higher accuracies than methods relying solely on either diversity or uncertainty. In particular, \emph{BADGE} attained the best performance among the baseline methods in the early stages of FashionMNIST learning. By incorporating these two AL modes into a single Gaussian process model, capturing the continuous learning behavior of $f$, our algorithm exhibited further significant improvements on CIFAR100, Caltech256, and Tiny ImageNet. For FashionMNIST, our algorithm's results were on par with the respective best-performing algorithms, namely \emph{WS} and \emph{BADGE}. The overall accuracies achieved by all algorithms in our experiments were considerably higher than the results reported in previous works~\cite{AZK20,SS18,YK19,SED19,KPK21}, primarily due to the use of stronger baseline learners.

\vspace{0.2cm}
\noindent\textbf{Contributions of Influence and Uncertainty:\;\;}
We evaluated two variations of our final algorithm, using only the $u^1$ utility (\emph{$u^1$-only}) and the $u^2$ utility (\emph{$u^2$-only}). Figure~\ref{f:ablation} shows the results. Our influence utility $u^1$ demonstrated greater effectiveness on CIFAR100 and in the early learning stages of FashionMNIST, where the predictions of the learner were less accurate. However, as the baseline learner provided more reliable confidence estimates in the later learning stages of FashionMNIST, its performance advantage over \emph{$u^2$-only} diminished. Conversely, our uncertainty-based utility $u^2$ exhibited the opposite behavior, performing better in later learning stages of FashionMNIST. By leveraging their complementary strengths, thereby trading influence and uncertainty, our final algorithm consistently outperformed \emph{$u^1$-only} and \emph{$u^2$-only}.  

\section{Conclusions}
\label{s:conclusions}
We have introduced a novel active learning algorithm that leverages a Gaussian process (GP) model as a surrogate for the baseline neural network learner, effectively identifying influential and difficult data points. By using the well-established Bayesian framework, our algorithm offers a rigorous approach to maximizing the information gain at each stage of the active learning process. To identify difficult points, an efficient GP surrogate is instantly updated each time a single label is provided with each new labeled instance. This allows us to faithfully simulate the continuous learning behavior of the baseline learner without the need for retraining. Consequently, we can avoid introducing additional mechanisms to promote label diversity, which often necessitate tuning separate hyperparameters. 

Our GP surrogate $\hat{f}$ may deviate from $f$. Empirically, we have observed that $\hat{f}$ faithfully captures the behavior of $f$ due to the use of the product kernel $k$ (Eq.~\ref{e:kernel}). Nonetheless, a theoretical analysis of the quality of $\hat{f}$ as a surrogate for $f$ and its impact on the resulting active learning performance would provide deeper insights into the utility of our algorithm. Future work should explore this.

\section*{Acknowledgments}
This work was supported by the National Research Foundation of Korea (NRF) grant (No. 2021R1A2C2012195) and the Institute of Information \& communications Technology Planning \& Evaluation (IITP) grants (No.2019-0-01906, Artificial Intelligence Graduate School Program, POSTECH, and 2020–0–01336, Artificial Intelligence Graduate School Program, UNIST), all funded by the Korea government (MSIT).

\bibliography{biblio}

\appendix

\section{Additional Results} 
\begin{figure*}[t]
\centering
\includegraphics[width=0.4835\linewidth]{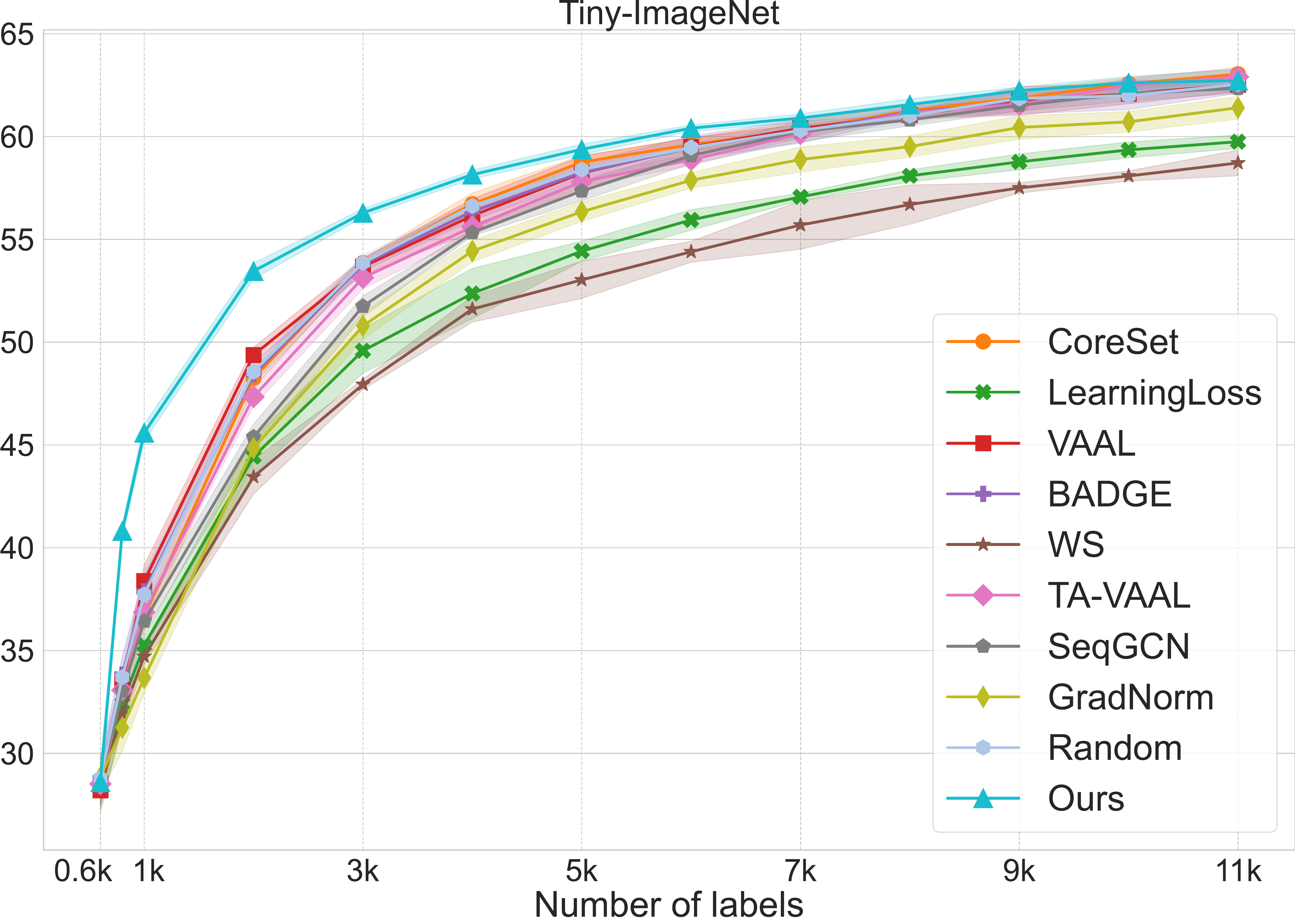}\;\;
\includegraphics[width=0.4835\linewidth]{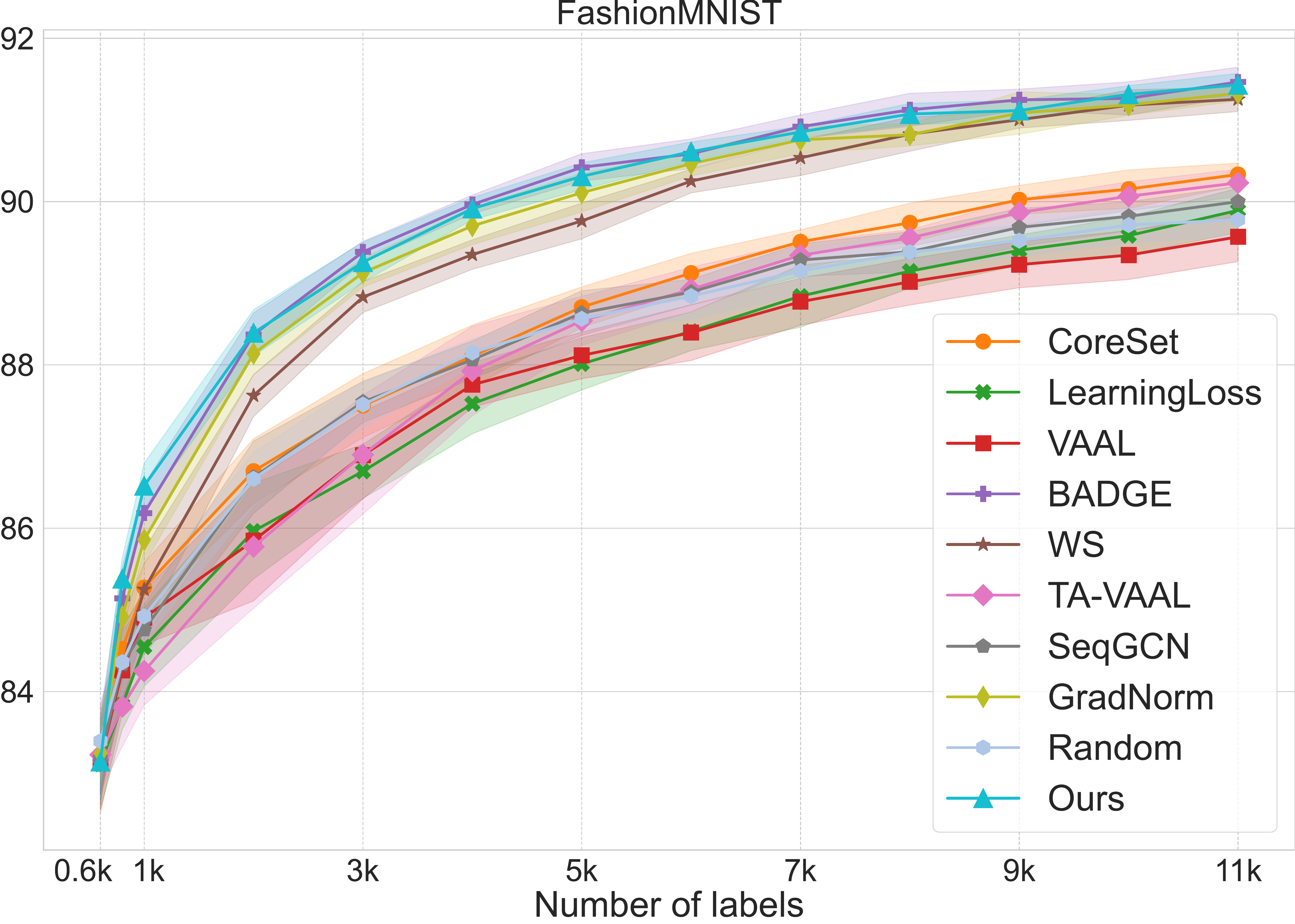}\\
\vspace{0.2cm}
\includegraphics[width=0.4835\linewidth]{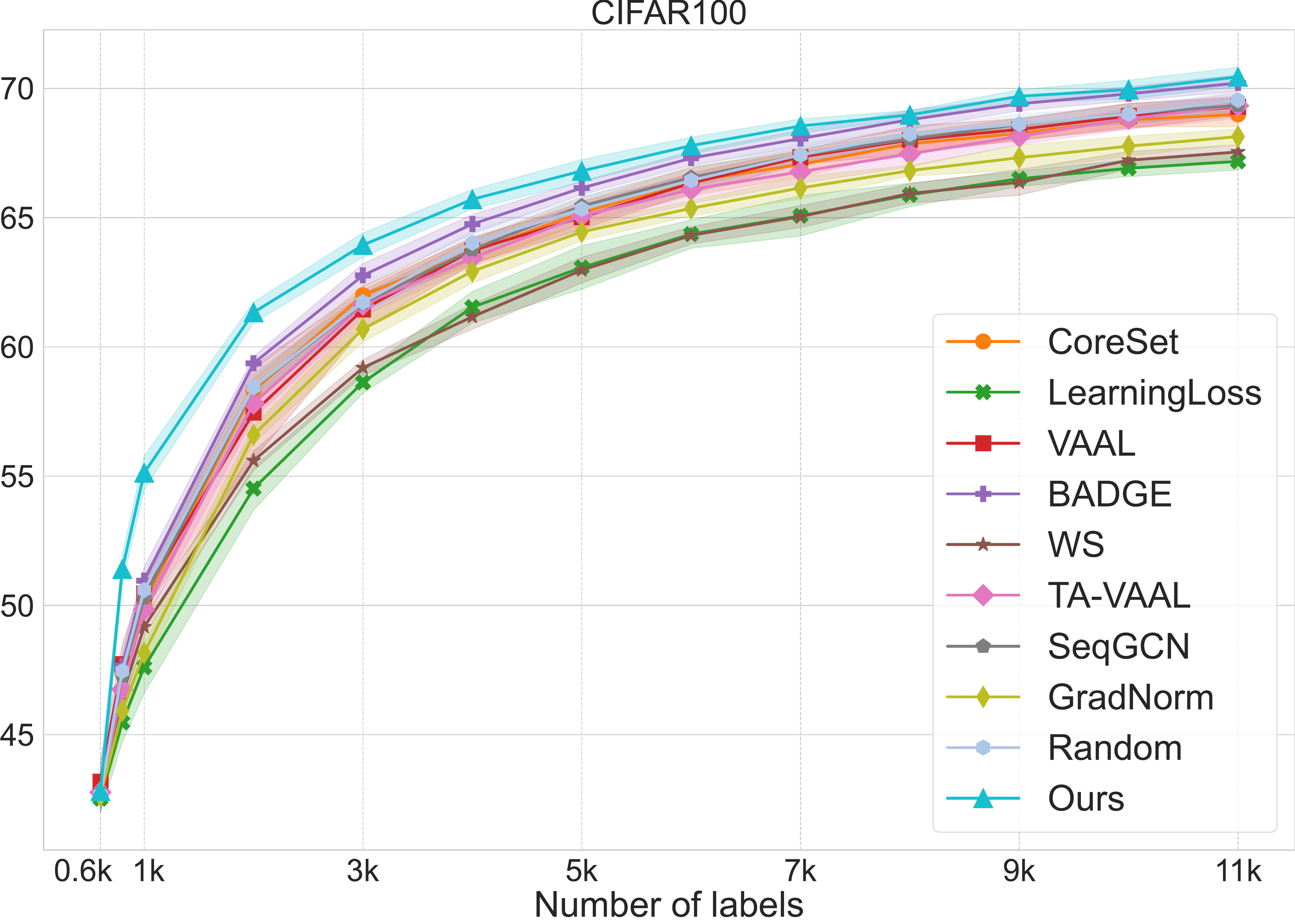}\;\;
\includegraphics[width=0.4835\linewidth]{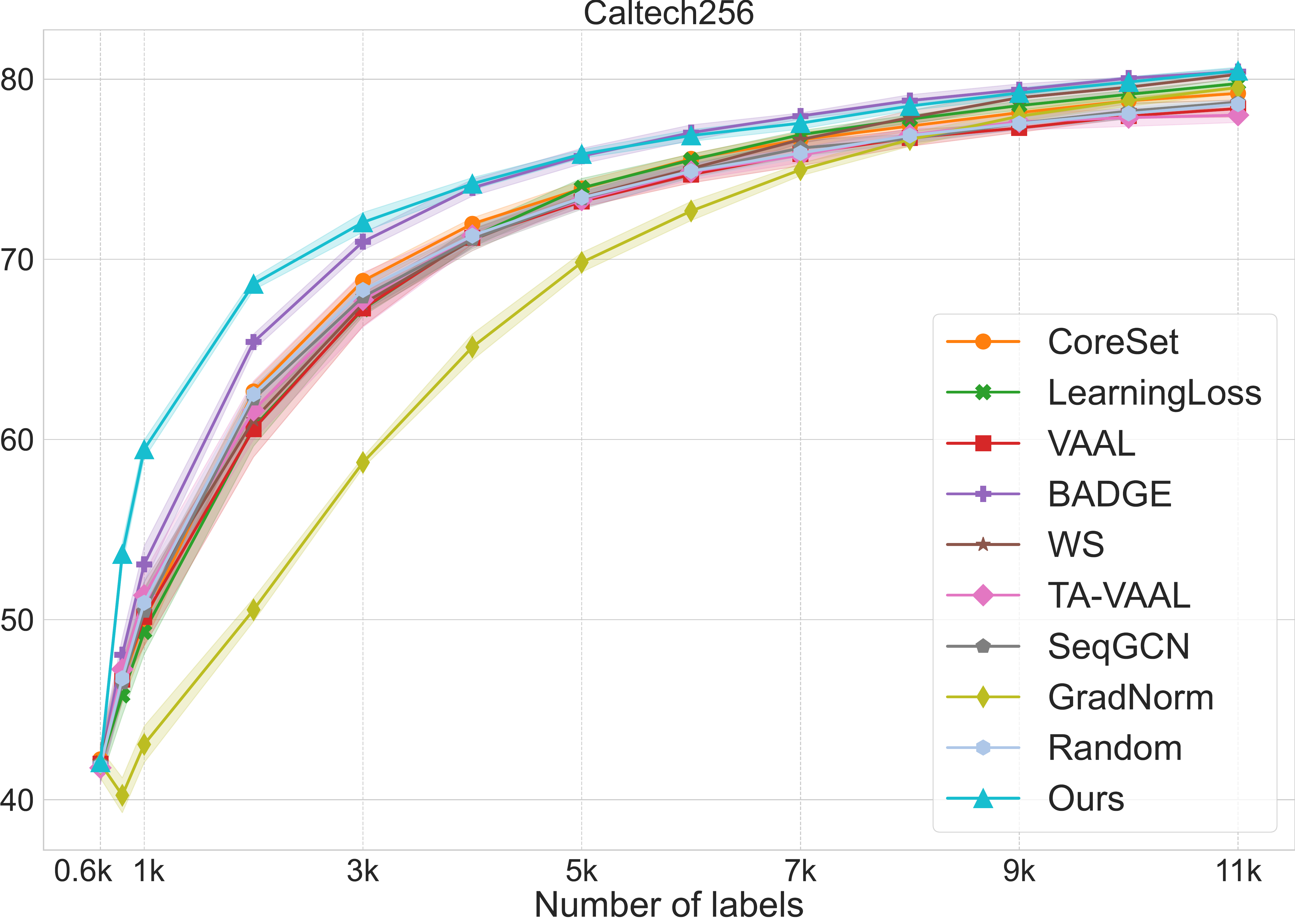}
\caption{Mean accuracy (in \%) across ten repeated experiments for different active learning algorithms including \emph{CoreSet}~\cite{SS18}, \emph{LearningLoss}~\cite{YK19}, \emph{BADGE}~\cite{AZK20}, \emph{SeqGCN}~\cite{CBK21}, \emph{GradNorm}~\cite{WLY22}, weight decay scheduling (\emph{WS})~\cite{YKK20}, \emph{VAAL}~\cite{SED19}, \emph{TA-VAAL}~\cite{KPK21}, \emph{Random}, and our algorithm, with random network initializations. The widths of the shaded regions represent twice the standard deviations.}
\label{f:supplresults}
\end{figure*}

In the main paper, we compared with \emph{CoreSet}, \emph{LearningLoss}, \emph{SeqGCN}, \emph{BADGE}, \emph{GradNorm}, and \emph{WS}. Here we also show the results obtained with random selection (\emph{Random}), \cite{SED19}'s variational adversarial active learning (\emph{VAAL}) and \cite{KPK21}'s task-aware VAAL-extension (\emph{TA-VAAL}). Figure~\ref{f:supplresults} demonstrates that our algorithm consistently exhibits enhanced performance.

\section{Varying Hyperparameters} 
Our hyperparameters were fixed across datasets, determined without leveraging any specific information about the problem or data at hand. The number of basis points, $K$, was selected with a focus on computational efficiency. We chose a small positive value for $\sigma^2$ to ensure the positive definiteness of the regularized kernel matrices ($\mbK+\sigma^2\mbI$ and its low-rank approximation). The selection of the kernel width parameter, $\sigma_\mbx$, is more intricate. $\sigma_\mbx$ was determined based on average distances between data points such that $\sigma_\mbx^2$ scales in a manner similar to the data variance. While tuning these hyperparameters per task and dataset can potentially enhance performance, it often necessitates additional validation labels. Obtaining such labels can be challenging in active learning scenarios where labeled data is scarce.

On CIFAR10, increasing the number of basis points $K$ (Eq.~3 in the main paper) slightly improved the performance until it reached 2,000 (up to approximately $0.06\%$, averaged across different numbers of labels), after which the performance saturated. We chose $K=500$ to balance computational complexity (which is quadratic in $K$). Modifying the original $\sigma_\mbx$ value (Eq.~1 in the main paper) by a factor in the range of $[0.3,8]$ had a negligible impact on performance. However, decreasing the magnification factor below 0.1 led to a rapid decrease in accuracy. Similar effects were observed when varying $\sigma_\mby$ (Eq.~1 in the main paper). The noise level $\sigma^2$ (Eq.~2 in the main paper) was set to a small value of $10^{-10}$, and varying it within the range of ${10^{-8},10^{-9},10^{-11},10^{-12}}$ did not result in any noticeable performance variation.

\begin{figure}[t]
\centering
\includegraphics[width=\columnwidth]{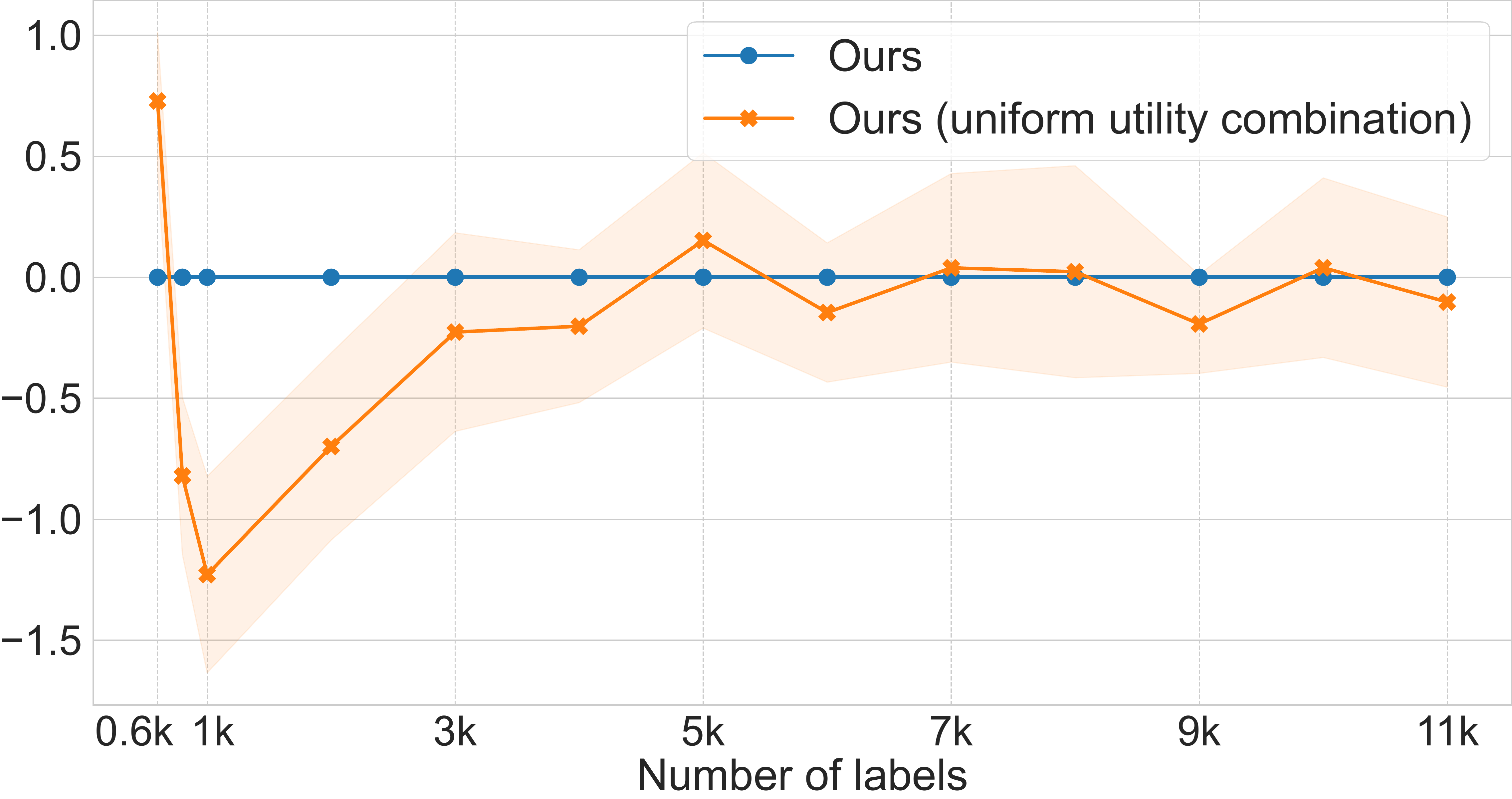}
\caption{The results of our final algorithm variation, `Ours (uniform utility combination)', employing a uniform combination of $u^1$ and $u^2$.}
\label{f:uniform}
\end{figure}

Our final algorithm combines the utility functions $u^1$ and $u^2$ via the following convex combination:
\begin{align}
u(i) = (1-P(f^t))\overline{u}^1(i)+P(f^t)\overline{u}^2(i).\nonumber
\end{align}
Here, $\overline{u}^1$ and $\overline{u}^2$ represent the normalized versions of $u^1$ and $u^2$ respectively, based on their respective standard deviations. $P(f^t) \in [0,1]$ is an estimate of the test accuracy. Figure~\ref{f:uniform} illustrates the results of an alternative version of our final algorithm, which employs a uniform combination of $u^1$ and $u^2$: $u=\frac{1}{2}u^1+\frac{1}{2}u^2$. These results exhibit a slight degradation, reinforcing the validity of our final design choice.

\begin{figure*}[t]
\centering
\includegraphics[width=0.48\linewidth]{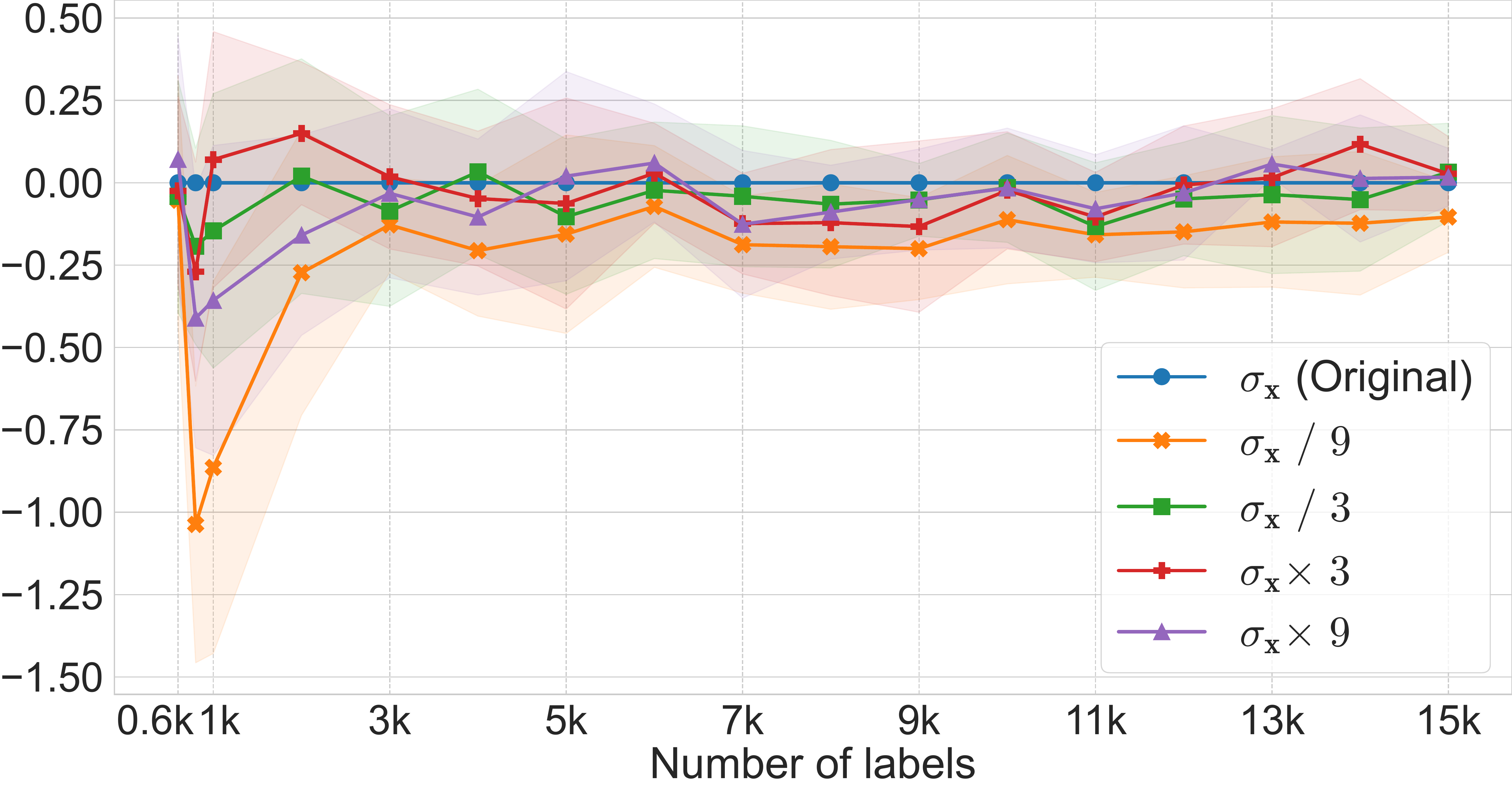}\;\;
\includegraphics[width=0.48\linewidth]{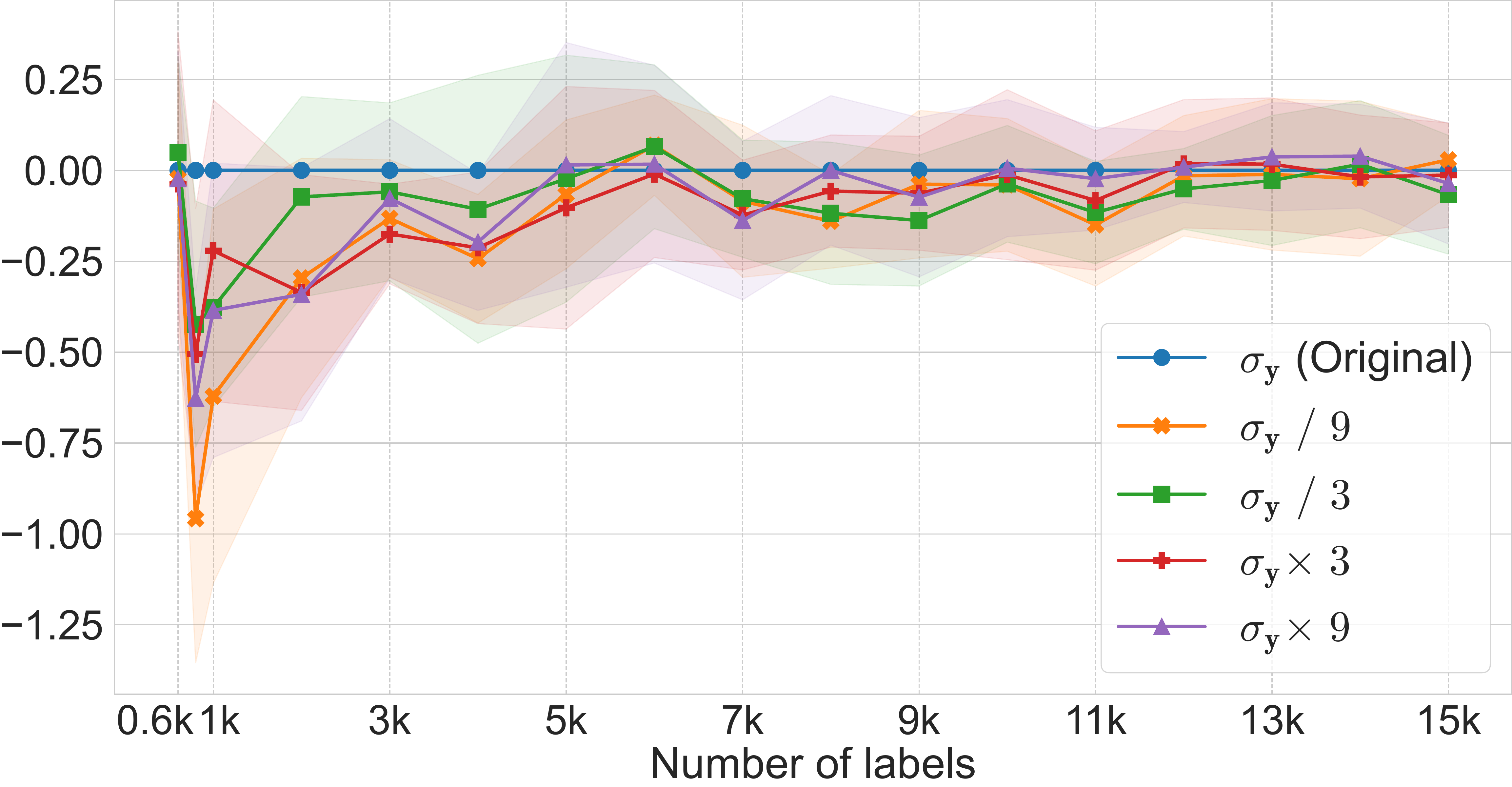}\\
\caption{The effect of varying hyperparameter $\sigma_\mbx$ (left) and $\sigma_\mby$ (right) values (CIFAR10 dataset). The y-axis represents the accuracy offset from our final algorithm.}
\label{f:hyperparameters}
\end{figure*}

\section{Gaussian Process Regression} 
Here, we provide a brief derivation of Gaussian process (GP) regression. For a comprehensive introduction, readers are referred to~\cite{RasWill06}. The notations used in this subsection differ slightly from those in the main paper.

Assume a set of training inputs $X={\mbx_1,\ldots,\mbx_N}\subset \calX$ and corresponding labels $Y={y_1,\ldots,y_N}\subset \R$ are provided. Our goal is to construct an estimate of an underlying latent regression function $f:\calX\mapsto\R$ based on $(X,Y)$. We adopt an independent and identically distributed (i.i.d.) Gaussian likelihood~\cite{RasWill06}:
\begin{align}
y_i=f(\mbx_i)+\epsilon, \text{ where } \epsilon\sim\calN(0,\sigma^2)
\label{e:latentgp}
\end{align}
and $\calN(\bm{\mu},\Sigma)$ is the Gaussian distribution with mean vector $\bm{\mu}$ and covariance matrix $\Sigma$. In Eq.~\ref{e:latentgp}, $\epsilon$ is one-dimensional, making $\bm{\mu}$ and $\Sigma$ one-dimensional as well. Subsequently, a zero-mean GP prior is imposed on $f$. For the training inputs $X$ and a given test input $\mbx_*$, this prior is expressed as:
\begin{align}
p(\mbf^*,\mbf) = \calN\left(\mathbf{0},\begin{bmatrix}
\mbK_{\mbf,\mbf} & \mbk_* \\
\mbk_*^\top & k(\mbx_*,\mbx_*)
\end{bmatrix}\right),
\label{e:jointdistr}
\end{align}
where the subscript $\mbf$ represents indexing across training data points:
\begin{align}
\mbf &= [f(\mbx_1),\ldots,f(\mbx_N)]^\top,\nonumber\\
\mbf_* &= f(\mbx_*),\nonumber\\
[\mbK_{\mbf,\mbf}]_{i,j}&= k(\mbx_i,\mbx_j)\nonumber\\
[\mbk_*]_i &= k(\mbx_i,\mbx_*).\nonumber
\end{align}
Any positive definite function can serve as the \emph{covariance} function $k$. We use the standard Gaussian kernel:
\begin{align}
k(\mbx,\mbx',\sigma_\mbx^2)=\exp\left(-\frac{\|\mbx-\mbx'\|^2}{\sigma_\mbx^2}\right).\nonumber
\end{align}
Combining Eqs.~\ref{e:latentgp} and \ref{e:jointdistr}, the joint distribution $p(\mbf_*,\mbx_*,Y,X)$ is obtained as a Gaussian random vector:
\begin{align}
p(\mbf_*,\mbx_*,Y,X) = \calN\left(\mathbf{0},\begin{bmatrix}
\mbK_{\mbf,\mbf}+\sigma^2\mbI & \mbk_* \\
\mbk_*^\top & k(\mbx_*,\mbx_*)
\end{bmatrix}\right).\nonumber
\end{align}
From this joint Gaussian distribution, the predictive distribution $p(\mbf_*|\mbx_*,Y,X)$ is constructed by conditioning $\mbf_*$ on the labels $Y$:
\begin{align}
p(\mbf_*|\mbx_*,Y,X) &= \calN(\mbk_*^\top\big(\mbK_{\mbf,\mbf}+\sigma^2\mbI)^{-1}\mby),\nonumber\\
&k(\mbx_*,\mbx_*)-\mbk_*^\top(\mbK_{\mbf,\mbf}+\sigma^2\mbI)^{-1}\mbk_*\big),
\end{align}
where $\mby=[y_1,\ldots,y_N]$.

\end{document}